\documentclass[times,twocolumn,final]{elsarticle}
\usepackage{medima-arxiv}
\usepackage{framed,multirow}
\usepackage{amsmath}
\usepackage{amssymb}
\usepackage{array}
\usepackage{latexsym}
\usepackage{url}
\usepackage[breaklinks]{hyperref}
\usepackage{romannum}
\usepackage{bm}
\usepackage{booktabs}
\usepackage[table]{xcolor}
\usepackage{subfig}
\definecolor{anti-flashwhite}{rgb}{0.95, 0.95, 0.96}
\definecolor{whitesmoke}{rgb}{0.94, 0.94, 0.94}
\definecolor{teagreen}{rgb}{0.82, 0.94, 0.75}
\definecolor{powderblue}{rgb}{0.69, 0.88, 0.9}
\definecolor{pastelblue}{rgb}{0.68, 0.78, 0.81}
\definecolor{lightskyblue}{rgb}{0.53, 0.81, 0.98}
\definecolor{turquoise}{cmyk}{0.65,0,0.1,0.3}
\definecolor{purple}{rgb}{0.65,0,0.65}
\definecolor{dark_green}{rgb}{0, 0.5, 0}
\definecolor{orange}{rgb}{0.8, 0.6, 0.2}
\definecolor{red}{rgb}{0.8, 0.2, 0.2}
\definecolor{darkred}{rgb}{0.6, 0.1, 0.05}
\definecolor{blueish}{rgb}{0.0, 0.3, .6}
\definecolor{light_gray}{rgb}{0.7, 0.7, .7}
\definecolor{pink}{rgb}{1, 0, 1}
\definecolor{greyblue}{rgb}{0.25, 0.25, 1}
\definecolor{light_cyan}{rgb}{0.88,1,1}
\hypersetup{pdfauthor={Yi Lin, Zhengjie Zhu, Kwang-Ting Cheng, Hao Chen}}

\newcommand{\eg}{\textit{e}.\textit{g}.}

\usepackage{pifont}
\newcommand{\myparagraph}[1]{\vspace{0.1em}\noindent\textbf{#1}}
\usepackage{arydshln}
\definecolor{newcolor}{rgb}{.8,.349,.1}
\journal{Medical Image Analysis}
\begin{document}
% -------------------------------------
\verso{Yi Lin, \textit{et~al.}}
% -------------------------------------
\begin{frontmatter}
% -------------------------------------
\title{Prompt-Guided Foundation Model Tuning for Pathology Image Classification}
% -------------------------------------
\author[1]{Yi \snm{Lin}\fnref{fn1}}
% -------------------------------------
\author[1]{Zhengjie \snm{Zhu}\fnref{fn1}}
\fntext[fn1]{Y. Lin and Z. Zhu contribute equally to this work.}
% -------------------------------------
\author[1,2]{Kwang-Ting \snm{Cheng}}
% -------------------------------------
\author[1,3,4,5]{Hao \snm{Chen}\corref{cor1}}
\cortext[cor1]{Corresponding author: Hao Chen (E-mail:~jhc@cse.ust.hk).}
% -------------------------------------
\address[1]{Department of Computer Science and Engineering, The Hong Kong University of Science and Technology, Hong Kong, China.}
\address[2]{Department of Electronic and Computer Engineering, The Hong Kong University of Science and Technology, Hong Kong, China.}
\address[3]{Department of Chemical and Biological Engineering, The Hong Kong University of Science and Technology, Hong Kong, China.}
\address[4]{HKUST Shenzhen-Hong Kong Collaborative Innovation Research Institute, Futian, Shenzhen, China.}
\address[5]{State Key Laboratory of Nervous System Disorders, The Hong Kong University of Science and Technology, Hong Kong, China
}
% -------------------------------------
\begin{abstract}
Foundation models have become pivotal in advancing computational pathology, particularly for whole slide image (WSI) classification. 
However, prevailing methodologies often rely on frozen, pre-trained models for feature extraction, overlooking the pronounced domain shift and task discrepancy between the pre-training and downstream tasks.
To address this challenge, we propose \texttt{PAMT}, a novel Prompt-guided Adaptive Model Transformation framework that enables precise adaptation of general foundation models to the distinct domain of histopathology.
To encapsulate the intricate distributions characteristic of histopathological data, we introduce Representative Patch Sampling (RPS) and Prototypical Visual Prompt (PVP), which reconstruct the input into compact yet highly informative representations. 
Further, to effectively bridge the domain gap, we incorporate Adaptive Model Transformation (AMT) via adapter modules within the feature extraction pipeline, facilitating the acquisition of domain-specific features by the foundation model.
We conduct rigorous evaluation across 14 publicly available datasets and demonstrate consistent, substantial improvements in classification accuracy. 
These results establish \texttt{PAMT} as a compelling new benchmark for pathology image classification and underscore the critical value of targeted model adaptation within computational pathology.
\fntext[fn2]{Source code is available at \url{https://github.com/hust-linyi/PAMT}.}
\end{abstract} 
% ----------------------------------------
\end{frontmatter}
% -----------------------------------------------------
\section{Introduction}
Whole slide image (WSI) is indispensable in modern histopathology, offering a detailed view crucial for accurate clinical disease diagnosis and research~\citep{3}. The advent of deep learning techniques has significantly transformed histopathology image analysis, promising improvements in both precision and efficiency~\citep{4}. 
Despite these advancements, deep-learning-based WSI classification presents notable challenges due to the massive size of WSIs, necessitating the division of WSIs into smaller patches for analysis~\citep{hou2016patch}. 
This patch-based approach, however, introduces labor-intensive and time-consuming annotation processes, limiting the effectiveness of supervised learning methods in this context~\citep{7}.

To circumvent these limitations, multiple instance learning (MIL) has become the preferred strategy for WSI analysis. In MIL, a WSI is treated as a bag of patches, and the entire bag is labeled based on the presence or absence of disease indicators among its constituent patches. However, this method requires effective downsampling and feature extraction strategies to manage the voluminous data, with the quality of patch features directly impacting MIL classification outcomes~\citep{ABMIL,li2021dual,DTFD,CLAM}. 
Existing MIL-based methods mainly leverage pre-trained feature extractors without fine-tuning due to the unaffordable computational cost of training the entire model on the large-scale WSI datasets~\citep{ABMIL,li2021dual,DTFD,CLAM}, ignoring the domain shift and task discrepancies between the pre-training task (\eg, ImageNet) and the downstream task (\eg, histopathology)~\citep{stacke2020measuring}.
To mitigate domain shift, recent efforts have developed pathology-specific foundation models pre-trained on large-scale histopathology datasets~\cite{lu2023visual}. These approaches range from self-supervised learning on images~\cite{li2021dual} to multi-modal learning combining images and text~\cite{li2023task}. 
However, computational constraints typically necessitate freezing these models during downstream WSI classification, limiting their adaptation to specific tasks and failing to fully leverage available WSI labels. 
While parameter-efficient fine-tuning (PEFT) techniques such as visual prompts~\cite{zhang2023prompt} and adapter tuning~\cite{zhu2023melo} have shown promise in natural image domains, their application to WSI classification remains largely unexplored.

In this work, we propose \texttt{PAMT}, a Prompt-guided Adaptive Model Transformation framework that enhances pathology image classification by ingeniously reprogramming both the input bags and intermediate features with the visual prompt and model adapter. 
For this purpose, we first introduce a Representative Patch Sampling (RPS) strategy to capture the most relevant and informative patches while discarding redundant ones, which extensively reduces the computational cost and enables efficient end-to-end training.
Additionally, to encapsulate the diverse spectrum of pathological data distributions, we further propose Prototypical Visual Prompt (PVP) module, where visual prompts are added to the sampled patches, enhancing the model's ability to interpret the wide array of tissue characteristics inherent to WSI.
To facilitate model adaptation, we introduce an Adaptive Model Transformation (AMT) technique to modify the pre-trained feature extractor with scalable adapters, allowing nuanced adaptation of ImageNet-trained architectures to the requirements of pathological image analysis.
The pre-trained model is frozen while training visual prompts, adapter blocks, and a lightweight MIL classifier, ensuring the intrinsic knowledge of the model is maintained.  

Our contributions are succinctly summarized as follows:

\begin{itemize}
    \item We propose PAMT, a comprehensive dual-level reprogramming framework specifically tailored for pathology image classification. By systematically addressing both input- and model-level adaptation, PAMT effectively narrows the domain gap and mitigates task discrepancy between pre-training and downstream histopathology tasks.
    \item We develop three synergistic components to overcome the unique challenges of WSI classification: (1) Representative Patch Sampling (RPS) efficiently handles gigapixel sparsity by selecting informative patches for end-to-end optimization; (2) Prototypical Visual Prompts (PVP) model extreme tissue heterogeneity through a novel cluster-based prompt assignment; and (3) Adaptive Model Transformation (AMT) enables task-specific feature learning via lightweight adapters in a frozen backbone.
    \item Extensive validation on 14 public datasets spanning multiple cancer types demonstrates that PAMT significantly enhances classification performance across varied architectures and foundation models, establishing a robust paradigm for targeted model adaptation within computational pathology.
\end{itemize}

% -----------------------------------------------------
\section{Related Work}
\begin{figure*}[!t]
	\centering
	\includegraphics[width=\textwidth]{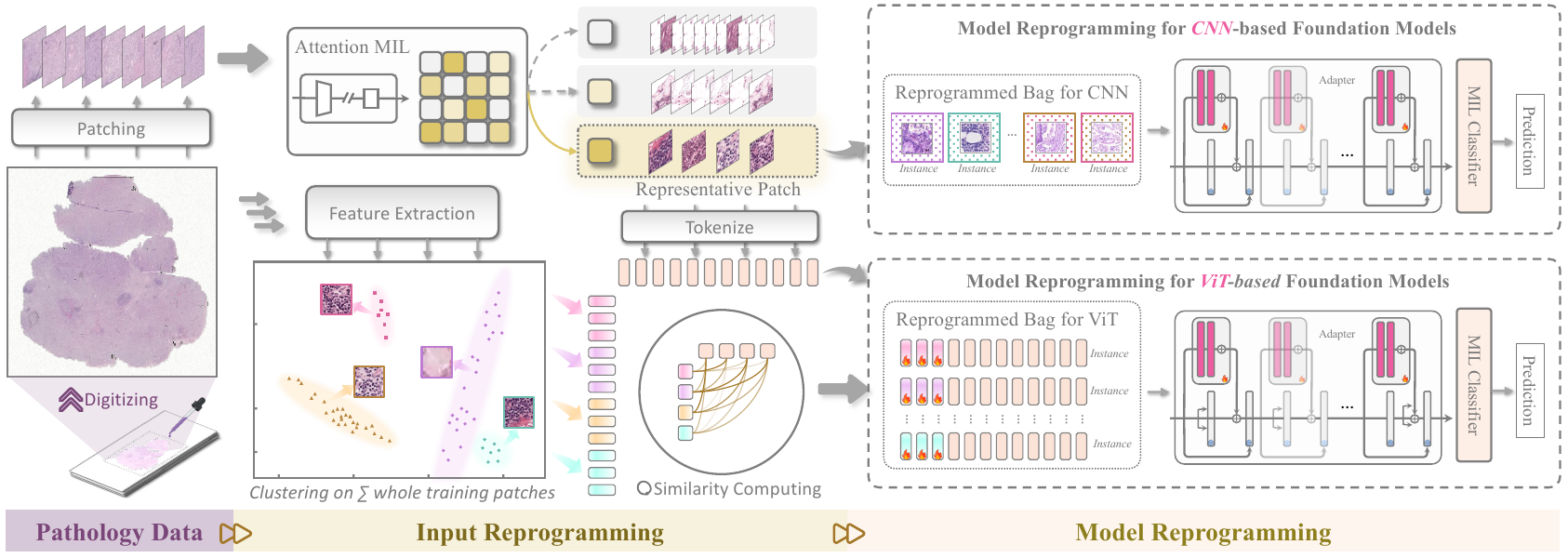}
	\caption{Overview of the proposed PAMT.} 
	\label{fig_overview}
\end{figure*}
\noindent\textbf{Multiple Instance Learning in CPATH.}
Multiple instance learning (MIL)-based approaches~\citep{ABMIL,li2021dual,DTFD,CLAM,lu2023visual,CLAM,shao2021transmil,li2024dynamic,lin2022insmix,lin2024bonus,zheng2022graph,huang2023conslide,yang2024mambamil,zhao2025ptcmil,zhao2024less} have gained popularity in computational pathology (CPATH), particularly in whole slide image (WSI) classification, due to their high efficacy in capturing complex spatial relationships and patterns within gigapixel-scale images. 
These methods typically adhere to a two-stage processing pipeline that begins with a feature extraction step, wherein deep neural network-based extractors are employed to derive features from individual image patches of the WSI.
Subsequently, an aggregation step is performed to obtain a comprehensive feature representation at the WSI level, which is then utilized by a lightweight classifier to predict the WSI category~\citep{ABMIL}.
Conventional MIL-based methods often utilize non-parametric aggregation functions, such as max and mean pooling, to summarize the features of image patches into a single WSI-level feature vector~\citep{CLAM}.
Afterwards, several studies have been conducted to enhance the aggregation process by incorporating attention mechanisms~\citep{ABMIL,li2021dual,huang2023conslide,shao2021transmil}, graph neural networks~\citep{zheng2022graph,li2024dynamic}, and bag-instance relationships~\citep{DTFD,lu2023visual,CLAM} to improve the classification performance of WSI.
In WSI classification, an effective feature extractor that generates representative features is crucial for accurate classification results. 
However, existing MIL-based methods for WSI classification mainly adopt a pre-trained feature extractor without fine-tuning, which results in sub-optimal performance~\citep{hu2022predicting}. 
This is due to the domain shift and task discrepancies between the pre-training task (\eg, ImageNet) and the downstream task (\eg, histopathology)~\citep{stacke2020measuring}. 
To narrow the domain gap, pathology-specific foundation models, such as PLIP~\citep{huang2023visual} and CONCH~\citep{lu2024visual}, have been proposed to pre-train large-scale models on histopathology datasets.
However, due to the limited availability of computational resources, these models are often frozen during the downstream WSI classification task, which may not fully leverage the potential of the pre-trained models~\citep{huang2023visual,lu2024visual}.
To address this issue, we introduce visual prompts for WSI classification, enabling smooth feature modulation from the upstream dataset to the downstream WSI classification.

\noindent\textbf{Foundation Models in CPATH.}
Foundation models have recently gained significant attention in the field of computational pathology (CPATH)~\citep{huang2023visual,lu2024visual,khan2025comprehensive,xiong2025survey}.
These models are typically large-scale pre-trained models that are trained on extensive datasets and can be adapted to various downstream tasks with minimal task-specific modifications.
The pioneering work adopted self-supervised learning to pre-train the models on large-scale histopathology images, such as SimCLR~\citep{chen2020simple} and DINO~\citep{ma2025generalizable}, to learn generalizable representations for histopathology images~\citep{li2021dual,chen2022scaling}.
Subsequently, vision-language pre-training methods, comprising vision and language encoders, have been widely adopted in CPATH to learn joint representations of histopathology images and associated text descriptions~\citep{huang2023visual,lu2024visual,chen2024towards,xu2024whole,vorontsov2024foundation,wang2024pathology}.
However, the majority of these foundation models are typically pre-trained on large-scale datasets and then frozen during downstream tasks, which may not fully leverage the potential of the pre-trained models~\citep{huang2023visual,lu2024visual}.
In this paper, we investigate the effectiveness of transfer learning in the context of pathology image classification by introducing visual prompts to adapt the pre-trained feature extractor for WSI classification, 
excavating the potential of the foundation models in CPATH.

\noindent\textbf{Parameter-Efficient Fine-Tuning.}
Parameter-efficient fine-tuning (PEFT) has recently emerged as a lightweight and efficient transfer learning paradigm in natural language processing (NLP) and has achieved remarkable success~\citep{jia2022visual}.
The fundamental idea behind PEFT is to freeze large-scale pre-trained models, such as BERT~\citep{devlin2018bert} and GPT-3~\citep{brown2020language}, that have been trained on vast datasets and use task-specific prompts to adapt them to diverse downstream tasks without updating any parameters~\citep{bahng2022exploring}.
Building on the NLP PEFT paradigm, several studies~\citep{luddecke2022image,jin2025lor,he2025rate,cox2024brainsegfounder} have proposed to extend PEFT to natural images in computer vision.
For example, L{\"u}ddecke et al.~\citep{luddecke2022image} used text and image prompts to adapt the frozen pre-trained CLIP model \citep{radford2021learning} to new image segmentation tasks.
In the context of computational pathology, the effectiveness of PEFT has been under-investigated. Prompt-MIL~\citep{zhang2023prompt} is the first work that introduces PEFT to MIL-based WSI classification, which uses visual prompts to adapt the pre-trained CLIP model for WSI classification. Qu et al.~\citep{qu2023rise} further extended the prompt learning to vision language pre-trained models for few-shot WSI classification. However, these existing visual prompting methodologies typically apply a single, uniform prompt to all patches within a bag, which overlooks the pronounced intra-class heterogeneity of WSIs. In this paper, our work focuses on scalable visual prompt learning for WSI classification by introducing Prototypical Visual Prompts (PVP). Unlike uniform prompting, PVP assigns cluster-specific learnable prompts to patches sharing similar histopathological features, empowering the model to interpret the wide array of tissue characteristics inherent to WSIs~\citep{chen2024towards}. This provides a more representative prompt learning paradigm for large-scale pre-trained models in histopathology image analysis.
% -----------------------------------------------------
\section{Method}
Figure~\ref{fig_overview} outlines our \texttt{PAMT} framework, structured in two critical phases: Input Reprogramming and Model Reprogramming. 
Initially, an attention-based MIL classifier is trained for Representative Patch Sampling (RPS). 
The input reprogramming phase then employs Prototypical Visual Prompts (PVP) to reprogram the input WSI bags. 
In the model reprogramming phase, we propose an adaptive model transformation (AMT) that fine-tunes visual prompts and adapter blocks in conjunction with the MIL classifier to ensure cohesive model adaptation. The nuances of each phase are elaborated in the following sections.

\subsection{Input Reprogramming}
\subsubsection{Attention-based MIL for WSI Classification}
To identify diagnostically informative patches for subsequent sampling, we first establish a baseline attention-based MIL classifier. 
For WSI classification via attention-based MIL, the conventional approach utilizes a frozen feature extractor $f(\cdot)$ pre-trained on ImageNet to extract all patch features. Then, all patch features in a WSI bag are aggregated to form the WSI feature using the attention mechanism, which assigns an attention score $\alpha_{k}$ for each patch $k$ through the MIL classifier.
Following \cite{CLAM}, we adopt a pruned ResNet-50 and integrate DTFD~\citep{DTFD}. 
This mechanism prioritizes patches more indicative of the WSI classification outcome. The WSI feature $\bm{F}$ is obtained by computing the attention-weighted average of all patch features in a WSI as~\cite{ABMIL}:
\begin{equation}
{\bm{F}} = \sum\limits_{k = 1}^{{K}} {{\alpha _{k}}{f(\bm{x}_{k})}},
\end{equation}  
where
\begin{equation}
{\alpha_k} = \dfrac{{\exp \left\{ {{{\bm{w}}^{\rm{T}}}\left( {\tanh \left( {{{\bm{V}}_1}{f(\bm{x}_{k})}} \right) \odot {\rm{sigmoid}}\left( {{{\bm{V}}_2}{f(\bm{x}_{k})}} \right)} \right)} \right\}}}{{\sum\limits_{j = 1}^K {\exp } \left\{ {{{\bm{w}}^{\rm{T}}}\left( {\tanh \left( {{{\bm{V}}_1}{f(\bm{x}_{j})}} \right) \odot {\rm{sigmoid}}\left( {{{\bm{V}}_2}{f(\bm{x}_{j})}} \right)} \right)} \right\}}},
\end{equation}  
where $\bm{w}$, $\bm{V}_1$ and $\bm{V}_2$ are learnable parameters in the MIL classifier, $\odot$ is the element-wise multiplication, and $\tanh(\cdot)$ and $\rm{sigmoid(\cdot)}$ denote the tanh and sigmoid activation functions, respectively.
Finally, the MIL classifier head $h(\cdot)$ predicts the label of WSI from the WSI feature $\bm{F}$, represented as:
\begin{equation}
\tilde{\bm y} = h(\bm{F}),
\end{equation}  
where $\tilde{\bm y}$ denotes the prediction of the WSI label. 
During training, we minimize the prediction error using the cross-entropy (CE) loss.

\subsubsection{Representative Patch Sampling}
Whole slide images contain substantial redundancy, with diagnostically significant regions often representing a small fraction of total tissue area~\citep{naik2020deep}. For instance, in Camelyon16~\citep{bejnordi2017diagnostic}  fewer than 10\% of patches in positive slides contain tumor tissue. Furthermore, fine-tuning foundation models on thousands of patches per WSI requires backpropagating gradients through the feature extractor, which is highly memory-intensive and computationally prohibitive, even with parameter-efficient techniques. This sparsity and corresponding computational bottleneck motivate a selective sampling strategy to reduce the computational burden while preserving discriminative information. We introduce representative patch sampling (RPS) to retain only the most informative patches and filter out redundant background or benign tissue. Specifically, the top-$K$ patches are selected from each original WSI bag $\mathcal{B}=\{B_{i}\}_{i=1}^{N}$, each containing $M_{i}$ patches $\{x_{ij}\}_{j=1}^{M_{i}}$. These patches are sampled based on their attention scores $\{\alpha_{ij}\}_{j=1}^{M_{i}}$ correlating highly with the WSI's overall diagnostic category,
yielding the subset:
\begin{equation}
\Phi(B_i) = {x_{ij} \mid \alpha_{ij} \in \text{top-$K$}(\{\alpha_{ij}\})},
\end{equation}
Setting $K=128$ patches per slide, ensures focus on patches most indicative of pathology, enabling efficient end-to-end model training.

\subsubsection{Prototypical Visual Prompts}
The selected representative patches exhibit substantial heterogeneity in tissue characteristics, including diverse cellular morphologies, stromal patterns, and staining variations. 
To adapt the pre-trained model to this heterogeneity while maintaining parameter efficiency, we propose prototypical visual prompts (PVP) that assign cluster-specific learnable prompts to patches sharing similar histopathological features. To capture the diverse tissue characteristics, we first perform an offline clustering (e.g., k-means) on all top-$K$ feature vectors extracted from the training set, defined as $\{f(x_{ij})|x_{ij}\in\Phi(\mathcal{B})\}$. This process forms $C$ clusters with the set of centers $\{\mu_{c}\}_{c=1}^{C}$, which serve as the anchors for visual prompts $\{p_{c}\}_{c=1}^{C}$. To maintain training stability, these cluster centers and the corresponding patch assignments are kept fixed during the subsequent end-to-end training phase. Each patch $x_{ij}\in\mathbb{R}^{H\times W}$ is assigned to its nearest fixed cluster and augmented with the corresponding trainable visual prompt:
\begin{equation}
x_{ij}' = x_{ij} + p_{\epsilon_{ij}},
\end{equation}
where
\begin{equation}
\epsilon_{ij} = \underset{c}{\text{argmin}} \, \|f(x_{ij}) - \mu_c\|_2,
\end{equation}
As depicted in Figure~\ref{fig_overview}, we implement distinct visual prompt designs for CNN and ViT architectures, tailored to their respective input processing paradigms. This approach empowers the model to adaptively recalibrate its attention, thereby enhancing its focus on diagnostically salient features.
Specifically, for the CNN architecture, we append the visual prompts as additional padding to the input patches, expanding the input size from $H \times W$ to $(H + 2p) \times (W + 2p)$.
Here, the number of parameters is $2Cp(H + W + 2p)$ for each padding prompt $p_{\epsilon_{ij}}$, with the padding size $p$. 
$C$ and $p$ are set to 4 and 10 based on parameter sensitivity analysis (Sec.~\ref{sec:hyperparameter}), guiding it toward a more nuanced interpretation of diverse pathological features within the patches.
For ViT, we follow \cite{zhang2023prompt} to insert learnable visual tokens into the input sequence of image embedding. The number of parameters is $C\times L\times D$ where $D$ is the dimension of each token, and $L$ is the number of prompt tokens for each patch. In our experiments, we set the number of prompt tokens $L=4$ to balance parameter efficiency with representation capacity.

Finally, the prompted patches are aggregated to form the reprogrammed bags $\Psi(\mathcal{B}) = \{{x'_{{ij}} \mid x_{ij} \in \Phi(B_i)}\}_{i=1}^{N}$, which serve as recalibrated input WSI bags, encapsulating the most discriminative features for robust prediction.
By focusing on the selected patches that are deemed highly informative and relevant, the recalibrated bags aim to enhance the classifier's sensitivity to diagnostically significant patterns.

\subsection{Model Reprogramming}
While input reprogramming enhances data representation, the frozen feature extractor still suffers from domain shift between natural images and histopathology. To address this, we introduce adaptive model transformation (AMT) that enables domain-specific feature learning while preserving pre-trained knowledge.
Input reprogramming and model reprogramming serve complementary roles: the former modifies patch representations at the input level, while the latter adapts internal feature representations within the frozen backbone. This dual-level adaptation enables end-to-end learning with parameter efficiency.
AMT retains the benefits of large-scale pre-training while tailoring the feature extractor to the unique demands of pathology images. 
Our framework augments the vision Transformer (ViT) or CNN architecture with adapter blocks positioned parallel to its foundational blocks. 
In ViT, we follow \cite{hu2022lora} to insert the low-rank adaptation (LoRA) module into last three self-attention layers of the Transformer block.
In CNNs, these adapters comprise a sequence of a global average pooling (GAP), followed by a two-layer multi-layer perceptron (MLP) and a sigmoid activation function, as depicted in Figure~\ref{fig_overview}. 
For an intermediate feature map $\bm{f}_{i-1}$ as the input for the $i$-th ResNet block, an adapter block generates a domain-adaptive vector $\bm{b}_i$.
This vector is then channel-wise multiplied with subsequent feature maps $g_i(\bm{f}_{i-1})$, facilitating a tailored feature adaptation, represented as:
\begin{equation}
\bm{b}_i = {\rm{sigmoid}}( \bm{W}_{i2} {\rm{ReLU}}(\bm{W}_{i1} {\rm{GAP}}(\bm{f}_{i-1}))),
\end{equation}
\begin{equation}
\bm{f}_{i} = g_i(\bm{f}_{i-1}) \odot \bm{b}_i,
\end{equation}
where $\text{ReLU}(\cdot)$ denotes rectified linear unit, $\bm{W}_{i1}$ and $\bm{W}_{i2}$ are the learnable parameters to be fine-tuned, and the parameters of $g_i(\cdot)$ remain frozen during training.
During the training process, only the parameters of visual prompts, adapter blocks and the lightweight MIL classifier are updated in an end-to-end manner, while the original pre-trained feature extractor is frozen. 
% -----------------------------------------------------
\section{Experiments}
\begin{table*}[!t]
\caption{Summary of the ten public patch-level histopathology datasets used for evaluation. All datasets are used for classification tasks. For the data split, '--' indicates that a dedicated set for that purpose was not specified in the official split.}
\label{tab:patch_dataset_summary}
\centering
\resizebox{\textwidth}{!}{
\renewcommand\arraystretch{1.2}
\setlength{\tabcolsep}{8pt}{
\begin{tabular}{@{\extracolsep{\fill}}l l c l r c}
\toprule
\textbf{Dataset} & \textbf{Primary Task} & \textbf{\# Classes} & \textbf{Image Size (px)} & \textbf{Total Patches} & \textbf{Data Split (Train / Val / Test)} \\
\midrule
BACH & Breast Tissue Type & 4 & $2048 \times 1536$\textsuperscript{a} & 400 & 320 / -- / 80 \\
BreakHis & Breast Cancer & 2 & Variable\textsuperscript{a} & 7,909 & 6,327 / -- / 1,582 \\
Chaoyang & Colon Tissue Type & 4 & Variable\textsuperscript{a} & 6,160 & 4,021 / -- / 2,139 \\
CRC-100K & Colorectal Tissue Type & 9 & $224 \times 224$ & 107,180 & 100,000 / -- / 7,180 \\
CRC-MSI & MSI Screening & 2 & $512 \times 512$ & 51,918 & 19,557 / -- / 32,361 \\
ESCA & Esophageal Subtyping & 11 & $256 \times 256$\textsuperscript{a} & 367,329 & 178,187 / -- / 189,142 \\
PanCancer-TCGA & Pan-Cancer Tissue Type & 32 & $256 \times 256$ & 271,170 & 217,368 / -- / 54,342 \\
PanCancer-TIL & TIL Detection & 2 & $100 \times 100$\textsuperscript{a} & 304,097 & 209,221 / 38,601 / 56,275 \\
PCAM & Metastasis Detection & 2 & $96 \times 96$\textsuperscript{a} & 327,680 & 262,144 / 32,768 / 32,768 \\
UniToPatho & Colorectal Polyp Grade & 6 & Not Specified & 9,536 & 6,270 / -- / 2,399 \\
\bottomrule
\end{tabular}}}
\begin{flushleft}
\textsuperscript{a}\footnotesize{Indicates that images are resized to a standard dimension (e.g., 224$\times$224 or 256$\times$256) for experimental consistency.}
\end{flushleft}
\end{table*}
\begin{table*}[!t]
\centering
\caption{Results of comparative experiments on Camelyon16, TCGA-NSCLC, and TUPAC16 datasets. In the table, the best results are in bold. The mean and standard deviation is reported (mean$_{\text{std}}$) under 3 times' experiments.}
\label{table1_new}%
\resizebox{\textwidth}{!}{
\renewcommand\arraystretch{1.2}
\setlength{\tabcolsep}{8pt}{
\begin{tabular}{lccc ccc ccc}
\toprule
& \multicolumn{3}{c}{Camelyon16} & \multicolumn{3}{c}{TCGA-NSCLC} & \multicolumn{3}{c}{TUPAC16}\\
\cmidrule(lr){2-4} \cmidrule(lr){5-7} \cmidrule(lr){8-10}
& AUC       & F1       & Acc      & AUC      & F1       & Acc   & AUC      & F1       & Acc   \\ 
\midrule
\rowcolor{whitesmoke} 
\multicolumn{10}{c}{ResNet-50}  \\
\midrule
 Mean-pooling & 0.590$_{0.018}$& 0.454$_{0.007}$& 0.677$_{0.011}$& 0.900$_{0.013}$& 0.834$_{0.011}$& 0.830$_{0.010}$& 0.629$_{0.011}$& 0.396$_{0.046}$& 0.507$_{0.017}$\\
 + PAMT & \textbf{0.810}$_{0.010}$& \textbf{0.714}$_{0.013}$& \textbf{0.803}$_{0.017}$& \textbf{0.925}$_{0.011}$& \textbf{0.874}$_{0.014}$& \textbf{0.871}$_{0.010}$& \textbf{0.651}$_{0.010}$& \textbf{0.428}$_{0.019}$& \textbf{0.525}$_{0.023}$\\
\hdashline
 Max-pooling  & 0.854$_{0.028}$& 0.754$_{0.119}$& 0.826$_{0.037}$& 0.906$_{0.021}$& 0.843$_{0.025}$& 0.832$_{0.027}$& 0.587$_{0.042}$& 0.384$_{0.037}$& 0.469$_{0.039}$\\
+ PAMT & \textbf{0.890}$_{0.015}$& \textbf{0.777}$_{0.016}$& \textbf{0.824}$_{0.025}$& \textbf{0.926}$_{0.013}$& \textbf{0.873}$_{0.011}$& \textbf{0.870}$_{0.013}$& \textbf{0.619}$_{0.013}$& \textbf{0.406}$_{0.044}$& \textbf{0.509}$_{0.012}$\\
\hdashline
 AB-MIL      & 0.874$_{0.023}$& 0.798$_{0.013}$& 0.861$_{0.009}$& 0.928$_{0.018}$& 0.878$_{0.014}$& 0.874$_{0.016}$& 0.647$_{0.006}$& 0.404$_{0.027}$& 0.514$_{0.025}$\\
 + PAMT      & \textbf{0.921}$_{0.011}$ & \textbf{0.828}$_{0.018}$& \textbf{0.866}$_{0.008}$& \textbf{0.931}$_{0.009}$& \textbf{0.884}$_{0.007}$& \textbf{0.879}$_{0.011}$& \textbf{0.669}$_{0.001}$& \textbf{0.503}$_{0.010}$& \textbf{0.549}$_{0.009}$\\
\hdashline
 CLAM-SB      & 0.902$_{0.012}$& 0.827$_{0.021}$& 0.871$_{0.019}$& 0.925$_{0.023}$& 0.870$_{0.026}$& 0.874$_{0.022}$& 0.641$_{0.014}$& 0.443$_{0.039}$& 0.499$_{0.057}$\\
 + PAMT      & \textbf{0.924}$_{0.009}$ & \textbf{0.832}$_{0.007}$& \textbf{0.881}$_{0.006}$& \textbf{0.933}$_{0.011}$& \textbf{0.891}$_{0.006}$& \textbf{0.888}$_{0.005}$& \textbf{0.666}$_{0.004}$& \textbf{0.495}$_{0.003}$& \textbf{0.541}$_{0.005}$\\
\hdashline
 DTFD-MIL     & 0.910$_{0.011}$& 0.835$_{0.029}$& 0.884$_{0.027}$& 0.930$_{0.019}$& 0.887$_{0.017}$& 0.884$_{0.021}$& 0.649$_{0.005}$& 0.412$_{0.015}$& 0.510$_{0.035}$\\
 + PAMT     & \textbf{0.929}$_{0.010}$& \textbf{0.859}$_{0.007}$& \textbf{0.897}$_{0.009}$& \textbf{0.942}$_{0.008}$& \textbf{0.895}$_{0.013}$& \textbf{0.893}$_{0.014}$& \textbf{0.673}$_{0.005}$& \textbf{0.459}$_{0.020}$& \textbf{0.531}$_{0.009}$\\ 
\hdashline
WiKG-MIL & 0.898$_{0.002}$& 0.814$_{0.012}$& 0.855$_{0.016}$& 0.924$_{0.004}$& 0.851$_{0.012}$& 0.852$_{0.011}$& 0.635$_{0.034}$& 0.419$_{0.050}$& 0.510$_{0.009}$\\
+ PAMT & \textbf{0.921}$_{0.004}$& \textbf{0.833}$_{0.011}$& \textbf{0.876}$_{0.008}$& \textbf{0.934}$_{0.001}$& \textbf{0.879}$_{0.002}$& \textbf{0.874}$_{0.002}$& \textbf{0.667}$_{0.016}$& \textbf{0.455}$_{0.045}$& \textbf{0.514}$_{0.046}$\\
\midrule
\rowcolor{whitesmoke} 
\multicolumn{10}{c}{ViT-b-32}  \\
\midrule
 Mean-pooling & 0.843$_{0.014}$& 0.740$_{0.014}$& 0.798$_{0.013}$& 0.939$_{0.015}$& 0.862$_{0.016}$& 0.861$_{0.015}$& 0.661$_{0.002}$& 0.442$_{0.016}$& 0.517$_{0.020}$\\
 + PAMT  & \textbf{0.924}$_{0.005}$& \textbf{0.841}$_{0.005}$& \textbf{0.876}$_{0.008}$& \textbf{0.969}$_{0.002}$& \textbf{0.909}$_{0.011}$& \textbf{0.909}$_{0.010}$& \textbf{0.671}$_{0.014}$& \textbf{0.460}$_{0.055}$& \textbf{0.528}$_{0.028}$\\
\hdashline
 Max-pooling  & 0.922$_{0.016}$& 0.824$_{0.021}$& 0.863$_{0.027}$& 0.948$_{0.007}$& 0.876$_{0.021}$& 0.875$_{0.020}$& 0.599$_{0.078}$& 0.395$_{0.040}$& 0.457$_{0.048}$\\
 + PAMT  & \textbf{0.940}$_{0.009}$& \textbf{0.869}$_{0.013}$& \textbf{0.904}$_{0.009}$& \textbf{0.965}$_{0.006}$& \textbf{0.903}$_{0.015}$& \textbf{0.899}$_{0.014}$& \textbf{0.633}$_{0.005}$& \textbf{0.401}$_{0.024}$& \textbf{0.515}$_{0.012}$\\
\hdashline
 AB-MIL       & 0.941$_{0.006}$& 0.839$_{0.004}$& 0.873$_{0.004}$& 0.951$_{0.015}$& 0.885$_{0.030}$& 0.880$_{0.025}$& 0.684$_{0.002}$& 0.497$_{0.007}$& 0.525$_{0.014}$\\
 + PAMT  & \textbf{0.958}$_{0.002}$& \textbf{0.903}$_{0.008}$& \textbf{0.930}$_{0.004}$& \textbf{0.974}$_{0.004}$& \textbf{0.918}$_{0.009}$& \textbf{0.914}$_{0.010}$& \textbf{0.698}$_{0.006}$& \textbf{0.510}$_{0.037}$& \textbf{0.557}$_{0.036}$\\
\hdashline
 CLAM-SB      & 0.941$_{0.011}$& 0.847$_{0.025}$& 0.889$_{0.016}$& 0.955$_{0.020}$& 0.885$_{0.025}$& 0.884$_{0.025}$& 0.674$_{0.003}$& 0.478$_{0.017}$& 0.511$_{0.026}$\\
 + PAMT  & \textbf{0.957}$_{0.006}$& \textbf{0.893}$_{0.019}$& \textbf{0.920}$_{0.016}$& \textbf{0.972}$_{0.007}$& \textbf{0.916}$_{0.014}$& \textbf{0.913}$_{0.017}$& \textbf{0.689}$_{0.004}$& \textbf{0.480}$_{0.057}$& \textbf{0.552}$_{0.024}$\\
\hdashline
 DTFD-MIL     & 0.946$_{0.009}$& 0.856$_{0.022}$& 0.891$_{0.013}$& 0.955$_{0.010}$& 0.885$_{0.026}$& 0.882$_{0.022}$& 0.652$_{0.002}$& 0.474$_{0.004}$& 0.500$_{0.024}$\\
 + PAMT  & \textbf{0.959}$_{0.006}$& \textbf{0.887}$_{0.009}$& \textbf{0.915}$_{0.008}$& \textbf{0.980}$_{0.001}$& \textbf{0.929}$_{0.009}$& \textbf{0.928}$_{0.007}$& \textbf{0.688}$_{0.016}$& \textbf{0.482}$_{0.052}$& \textbf{0.549}$_{0.041}$\\
\hdashline
WiKG-MIL & 0.942$_{0.010}$& 0.867$_{0.021}$& 0.897$_{0.018}$& 0.957$_{0.008}$& 0.892$_{0.012}$& 0.888$_{0.012}$& 0.660$_{0.007}$& 0.465$_{0.047}$& 0.523$_{0.044}$\\
+ PAMT & \textbf{0.954}$_{0.009}$& \textbf{0.896}$_{0.002}$& \textbf{0.922}$_{0.000}$& \textbf{0.966}$_{0.002}$& \textbf{0.905}$_{0.007}$& \textbf{0.902}$_{0.008}$& \textbf{0.685}$_{0.002}$& \textbf{0.472}$_{0.011}$& \textbf{0.541}$_{0.009}$\\
\bottomrule
\end{tabular}}
}
\end{table*}
\subsection{Datasets}
We evaluate our method on two pathology image classification settings: WSI-level disease classification and patch-level tissue classification.
For WSI-level classification, we conduct experiments on three public WSI benchmarks, including Camelyon16~\citep{bejnordi2017diagnostic}, TCGA-NSCLC~\citep{newman2015robust}, and TUPAC16~\citep{veta2019predicting}.
For patch-level classification, we conduct experiments on ten public patch classification benchmarks, including BACH~\citep{aresta2019bach}, BreakHis~\citep{spanhol2015dataset}, CRC-100K~\citep{kather2019predicting}, CRC-MSI~\citep{kather2019deep}, ESCA~\citep{tolkach2023artificial}, PCAM~\citep{veeling2018rotation}, PanCancer-TCGA~\citep{komura2022universal}, PanCancer-TIL~\citep{abousamra2022deep}, UniToPatho~\citep{barbano2021unitopatho}, and Chaoyang~\citep{zhu2021hard}.
We provide a brief introduction to each dataset as follows:
\begin{itemize}
    \item \textbf{Camelyon16}~\citep{bejnordi2017diagnostic} consists of 399 H\&E WSI from breast cancer screening, with two classes: normal and tumor.
    After preprocessing, we obtain a total of 4,610,687 patches at $20\times$ magnification, with an average of 11,556 patches per slide.
    We use the official 129 test set and split the official 270 training set into training and validation sets with a ratio of $9\colon1$.
    \item \textbf{TCGA-NSCLC}~\citep{newman2015robust} contains two subtypes of lung cancer: Lung Adenocarcinoma (LUAD) and Lung Squamous Cell Carcinoma (LUSC).
    After preprocessing, we obtain a total of 3,252,431 patches, with an average of 3,089 patches per WSI.
    We split the dataset into training, validation, and testing sets at the slide level, with a ratio of 65:10:25.
    \item \textbf{TUPAC16}~\citep{veta2019predicting} contains 500 WSIs for training and 321 WSIs for testing, all derived from breast cancer histopathology images. 
    \item \textbf{Camelyon17}~\citep{bandi2018detection} served as an external validation set for the Camelyon16 dataset. This dataset consists of 500 WSIs from five medical centers in the Netherlands. For binary WSI classification, we group slides with any type of metastasis as tumor and the remaining slides as normal.
    \item \textbf{Patch-level Classification Datasets}. We conduct comprehensive experiments on 10 public patch classification benchmarks, including various tissue types and disease states. We provide a detailed description in Table~\ref{tab:patch_dataset_summary}.
\end{itemize}

We adhere to the official data splits for training, validation, and testing as provided by the original authors or the referenced experimental setup. A key preprocessing step for fair comparison is resizing all input images to a consistent resolution (e.g., 224$\times$224 or 256$\times$256 pixels), as detailed in Table~\ref{tab:patch_dataset_summary}.

\subsection{Implementation Details}
\label{Implementation}
The performance is measured by Accuracy (Acc), F1 score (F1), and Area Under the Curve (AUC). 
In the representative patch sampling (RPS) process, we adopt the DTFD model~\citep{DTFD} as the attention-based MIL classifier.
The Adam optimizer \citep{kingma2014adam} was employed for the adapter blocks and classifier with a learning rate and weight decay of 1e-4 over 100 epochs. 
Following~\citep{bahng2022exploring}, visual prompts were optimized using SGD at a starting learning rate of 40, with the learning rate decayed according to a cosine schedule. 
These computations were facilitated by an NVIDIA GeForce RTX 4090 GPU.

\subsection{Comparison with State-of-the-Art Methods}
\subsubsection{WSI-level Disease Classification}
We compare our method with state-of-the-art methods on the Camelyon16, TCGA-NSCLC, and TUPAC16 datasets, including Mean-pooling, Max-pooling, AB-MIL~\citep{ABMIL}, CLAM-SB~\citep{CLAM}, DTFD-MIL~\citep{DTFD}, and WiKG-MIL~\citep{li2024dynamic}.
We evaluate the performance of our method with two popular backbones: CNN and Transformer.
For the CNN backbone, we use the ResNet-50 model pre-trained on ImageNet.
For the Transformer backbone, we use the ViT-b-32~\citep{dosovitskiyimage} pre-trained by PLIP~\citep{huang2023visual}.
To ensure a fair comparison, we re-implement all baseline methods with the same backbone and training strategy as our method. Specifically, for all baseline implementations, including Mean-pooling, Max-pooling, AB-MIL, CLAM-SB, DTFD-MIL, and WiKG-MIL, the pre-trained feature extractor (ResNet-50 or ViT-b-32) is completely frozen, and only the learnable parameters within the MIL aggregator and the final classification head are updated during training. For configurations incorporating our framework (denoted as ``+ PAMT"), only the parameters of visual prompts, adapter blocks and the lightweight MIL classifier are updated in an end-to-end manner, while the original pre-trained feature extractor is frozen. All experiments are repeated three times with different random seeds, and we report the mean and standard deviation of the results.

The results are shown in Table~\ref{table1_new}. With the ResNet-50 backbone, our method significantly improved the performance of the baseline methods by 22.0\%, 3.6\%, 4.7\%, 2.2\%, 1.9\%, and 2.3\% in AUC on the Camelyon16 dataset, respectively.
In terms of F1 score, our method achieved 0.714, 0.777, 0.828, 0.832, 0.859, and 0.833, improving the baseline methods by 26.0\%, 2.3\%, 3.0\%, 0.5\%, 2.4\%, and 1.9\%, respectively.
With the ViT-b-32 backbone, our method achieved 0.924, 0.940, 0.958, 0.957, 0.959, and 0.954 in AUC on the Camelyon16 dataset, improving the baseline methods by 8.1\%, 1.8\%, 1.7\%, 1.6\%, 1.3\%, and 1.2\%, respectively.
While in terms of F1 score, our method achieved 0.841, 0.869, 0.903, 0.893, 0.887, and 0.896, improving the baseline methods by 10.1\%, 4.5\%, 6.4\%, 4.6\%, 2.9\%, and 2.9\%, respectively.
The same trend is observed in the TCGA-NSCLC, and TUPAC16 datasets.
Notably, our method substantially improves the performance of both mean and max pooling strategies across datasets, indicating that enhanced feature quality enables even simple aggregation strategies to achieve competitive classification outcomes.

\subsubsection{Patch-level Tissue Classification}
\begin{table*}[!t]
\centering
\caption{Results of patch-level classification task on 10 public datasets.}
\label{tab:patch_cla}%
\resizebox{\textwidth}{!}{
\setlength{\tabcolsep}{8pt}
\renewcommand\arraystretch{1.2}
\begin{tabular}{l*{10}{c}}
\toprule
Dataset & FT & LP & VPT-s & VPT-d & LoRA & AdaLoRA & IA\textsuperscript{3} & AdaptF & SSF & Ours \\
\midrule
BACH & \textbf{0.972} & 0.956 & 0.870 & 0.816 & 0.800 & 0.800 & 0.588 & 0.796 & 0.625 & \underline{0.968} \\
BreakHis & \underline{0.998} & 0.989 & 0.983 & 0.984 & 0.995 & 0.995 & 0.956 & 0.994 & 0.969 & \textbf{0.998} \\
Chaoyang & \underline{0.946} & 0.938 & 0.912 & 0.912 & 0.945 & 0.945 & 0.905 & 0.937 & 0.914 & \textbf{0.950} \\
CRC-MSI & \textbf{0.729} & 0.693 & 0.681 & 0.687 & 0.705 & 0.705 & 0.701 & 0.707 & 0.694 & \underline{0.713} \\
CRC-100K & 0.978 & 0.989 & \underline{0.997} & \textbf{0.997} & 0.996 & 0.996 & 0.997 & 0.997 & 0.997 & 0.981 \\
ESCA & 0.895 & 0.889 & 0.975 & 0.976 & \textbf{0.982} & \underline{0.982} & 0.972 & 0.978 & 0.975 & 0.896 \\
PCAM & 0.951 & 0.945 & 0.957 & 0.958 & 0.961 & \underline{0.961} & 0.957 & \textbf{0.962} & 0.958 & 0.954 \\
PanCancer-TIL & \textbf{0.959} & 0.936 & 0.921 & 0.924 & 0.943 & 0.943 & 0.916 & 0.938 & 0.929 & \underline{0.955} \\
PanCancer-TCGA & \textbf{0.999} & 0.971 & 0.973 & 0.973 & 0.987 & 0.987 & 0.970 & 0.985 & 0.978 & \underline{0.998} \\
UniToPatho & \textbf{0.831} & 0.821 & 0.810 & 0.814 & 0.829 & \underline{0.829} & 0.809 & 0.827 & 0.815 & 0.826 \\
\midrule
Average & \textbf{0.926} & 0.913 & 0.908 & 0.904 & 0.914 & 0.914 & 0.877 & 0.912 & 0.885 & \underline{0.924} \\
\bottomrule
\end{tabular}
}
\end{table*}
While our framework is designed for WSI classification, the core adaptation mechanisms (PVP and AMT) can be applied to patch-level classification by bypassing RPS and replacing the MIL aggregator with a standard classifier. 
We compare our method with widely adopted fine-tuning paradigms: full-tuning, linear probing, and state-of-the-art parameter-efficient fine-tuning (PEFT) methods, including Visual Prompt Tuning (VPT)~\citep{jia2022visual}, LoRA~\citep{hu2021lora}, AdaLoRA~\citep{zhang2023adaptive}, IA\textsuperscript{3}~\citep{liu2022fewshot}, AdaptFormer~\citep{chen2022adaptformer}, and SSF~\citep{lian2022scaling}.
Experiments are conducted on ten public patch classification benchmarks, as detailed in Table~\ref{tab:patch_dataset_summary}.
Instead of using the representative patch sampling strategy for WSI-level classification, we directly utilize all patches in each dataset for patch-level classification. 
The results are shown in Table~\ref{tab:patch_cla}.
The empirical results reveal that our methodology attains performance metrics commensurate with the full-tuning strategy, while achieving a substantial reduction in the number of trainable parameters. 
Specifically, across ten benchmark datasets, our dual-level adaptation achieves an average AUC of 0.924, which significantly outperforms the linear probe baseline (0.913) and single-perspective PEFT methods such as VPT (0.908) and IA\textsuperscript{3} (0.877). 
A key advantage of our method is the synergistic integration of both data-centric and model-centric adaptations. While existing PEFT methods typically focus exclusively on either modifying model weights (e.g., LoRA, Adapters) or adjusting input representations (e.g., VPT), our framework couples prototypical visual prompts (PVP) with adaptive model transformation (AMT). This enables more comprehensive feature recalibration across diverse tissue types. 
Consequently, while top-performing structural PEFT methods like LoRA and AdaLoRA achieve an average AUC of 0.914, our framework consistently demonstrates superior capability, elevating the average AUC by an absolute margin of 1.0\% over these strong baselines. This performance closely parallels the full-tuning upper bound (0.926).
These findings underscore that approaches focusing exclusively on the MIL classifier are suboptimal for pathological image classification, presumably due to the pronounced domain disparity between pre-trained and target datasets. 
We further conduct statistical validation via the Wilcoxon signed-rank test~\citep{woolson2007wilcoxon} to compare our method with the full-tuning and linear probe baselines.
Compared to the full-tuning baseline, our method yields a p-value of 0.057, 0.572, and 0.474 for AUC, F1 score, and accuracy, respectively, indicating no significant difference between the two methods.
Compared to the linear probe baseline, our method consistently yields p-values below 0.0001 across all three metrics, indicating a statistically significant improvement over the linear probe approach.

\subsection{Ablation Study}
\subsubsection{Effectiveness of Our Method on Foundation Models.}
\begin{table}[!t]
\centering
\caption{Results of different foundation models under the DTFD-MIL framework on TUPAC16. Our PAMT module consistently improves performance. Best results in each pair are in bold. The mean and standard deviation are reported (mean$_{\text{std}}$).}
\label{tab:foundation_dtfd_tupac}
\renewcommand\arraystretch{1.2}
\setlength{\tabcolsep}{6pt}{ 
\begin{tabular}{l ccc}
\toprule
\textbf{Foundation Model} & AUC       & F1        & Acc       \\ 
\midrule
ResNet-50~(\citeyear{he2016deep}) & 0.653$_{0.002}$ & 0.460$_{0.028}$ & 0.486$_{0.022}$ \\
+ PAMT                     & \textbf{0.675}$_{0.021}$ & \textbf{0.485}$_{0.016}$ & \textbf{0.515}$_{0.038}$ \\
\hdashline
PLIP~(\citeyear{huang2023visual})                      & 0.652$_{0.002}$ & 0.471$_{0.004}$ & 0.515$_{0.024}$ \\
+ PAMT                    & \textbf{0.688}$_{0.016}$ & \textbf{0.482}$_{0.052}$ & \textbf{0.549}$_{0.041}$ \\
\hdashline
CONCH~(\citeyear{lu2024visual})                     & 0.684$_{0.001}$ & 0.497$_{0.023}$ & 0.547$_{0.020}$ \\
+ PAMT                    & \textbf{0.696}$_{0.004}$ & \textbf{0.528}$_{0.023}$ & \textbf{0.579}$_{0.012}$ \\
\hdashline
Prov-GigaPath~(\citeyear{xu2024whole})             & 0.656$_{0.005}$ & 0.504$_{0.022}$ & 0.520$_{0.042}$ \\
+ PAMT                    & \textbf{0.688}$_{0.001}$ & \textbf{0.543}$_{0.005}$ & \textbf{0.584}$_{0.014}$ \\
\hdashline
UNI~(\citeyear{chen2024towards})                       & 0.662$_{0.011}$ & 0.452$_{0.019}$ & 0.485$_{0.033}$ \\
+ PAMT                    & \textbf{0.693}$_{0.006}$ & \textbf{0.508}$_{0.016}$ & \textbf{0.544}$_{0.021}$ \\
\hdashline
Virchow~(\citeyear{vorontsov2024foundation})                   & 0.673$_{0.003}$ & 0.497$_{0.006}$ & 0.528$_{0.021}$ \\
+ PAMT                    & \textbf{0.690}$_{0.007}$ & \textbf{0.548}$_{0.018}$ & \textbf{0.584}$_{0.021}$ \\
\bottomrule
\end{tabular}}
\end{table}
We performed an ablation study on the TUPAC16 dataset to assess the efficacy of our proposed method when applied to various foundation models, including ResNet-50~\citep{he2016deep} pretrained on ImageNet-22K~\citep{deng2009imagenet}, as well as pathology foundation models including PLIP~\citep{huang2023visual}, CONCH~\citep{lu2024visual}, Prov-GigaPath~\citep{xu2024whole}, UNI~\citep{chen2024towards}, and Virchow~\citep{vorontsov2024foundation}.
As summarized in Table~\ref{tab:foundation_dtfd_tupac}, our approach confers substantial performance gains across all evaluated foundation models.
In particular, AUC scores are elevated by 2.2\%, 3.6\%, 1.2\%, 3.2\%, 3.1\%, and 1.7\% for ResNet-50, PLIP, CONCH, Prov-GigaPath, UNI, and Virchow, respectively.
Comparable enhancements are observed in both F1 score and accuracy, underscoring the generalizability of our method in bolstering the classification performance of diverse foundation models on pathology image analysis tasks.
It is noteworthy that foundation models pretrained specifically on pathological image data tend to surpass ResNet-50 pretrained on ImageNet-22K, thereby illustrating the merits of domain-adaptive pre-training.
Nevertheless, a considerable gap persists between the pre-training domain and the downstream task, constraining the attainable performance of these models.
Our results accentuate the pivotal role of the proposed method in bridging this domain/task discrepancy, further potentiating the utility of foundation models for pathology image classification tasks.
\begin{figure*}[thbp]
\centering
\subfloat[]{
\includegraphics[width=0.28\textwidth]{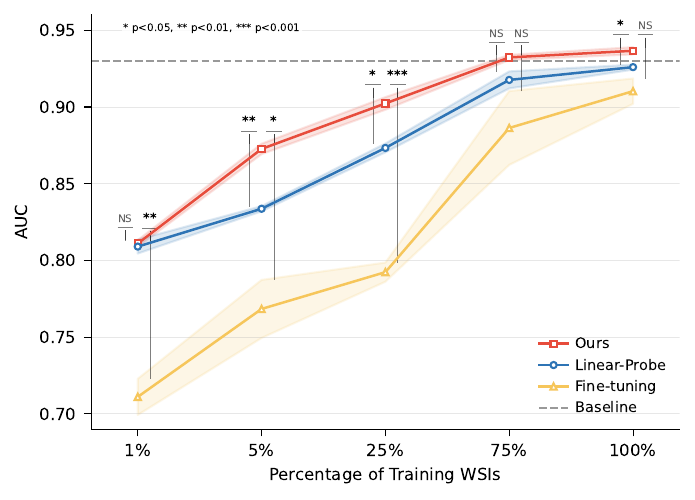}
\label{fig:wsi_num}
}
\subfloat[]{
\includegraphics[width=0.23\textwidth]{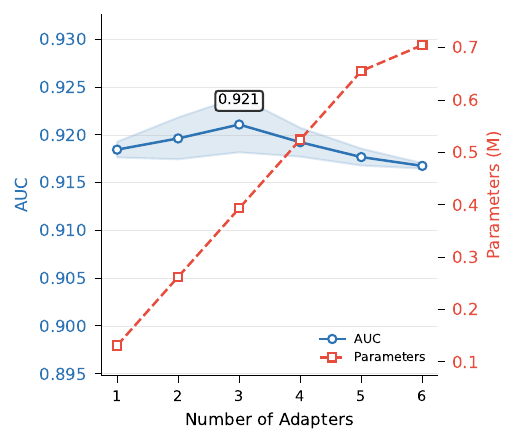}
\label{fig:adapter_num}
}
\subfloat[]{
\includegraphics[width=0.23\textwidth]{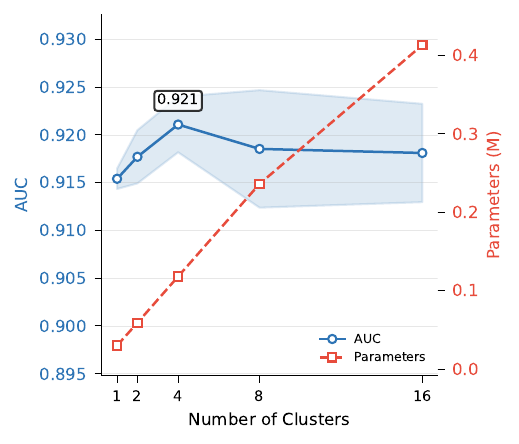}
\label{fig:cluster_num}
}
\subfloat[]{
\includegraphics[width=0.23\textwidth]{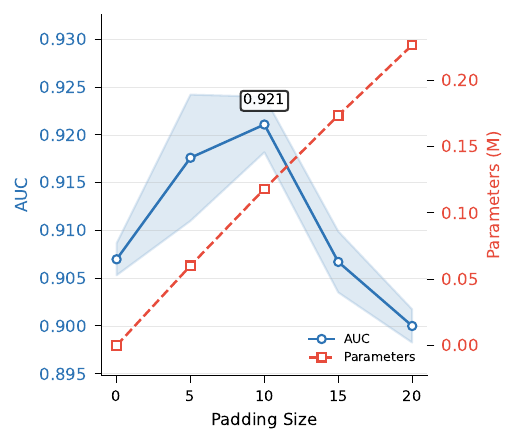}
\label{fig:padding_size}
}
\caption{Ablation Studies. (a) Different training strategies with varying proportions of training WSIs using the DTFD model on TCGA-NSCLC dataset; Parameter sensitivity analysis of (b) The number of adapters; (c) The number of clusters; and (d) The padding size.}
\end{figure*}

\begin{figure*}[thbp]
    \centering
    \includegraphics[width=0.95\textwidth]{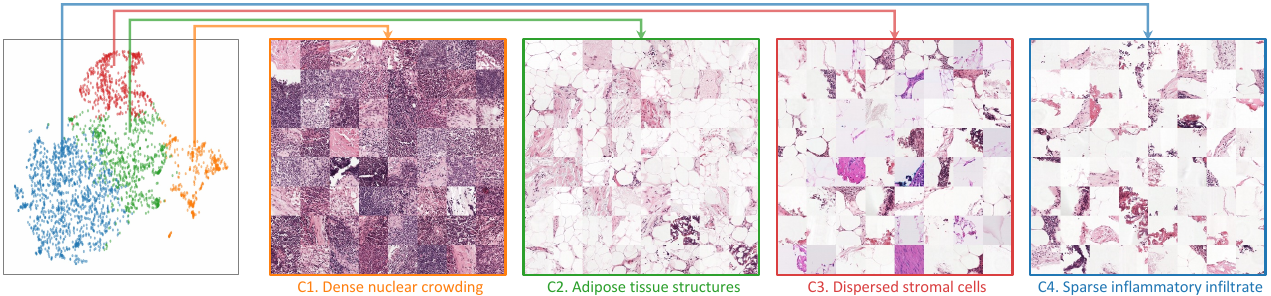}
    \caption{Visualization of cluster centers for input reprogramming. Four distinct clusters are showcased, representing varied histopathological features.}
    \label{fig_ablation_visual}
\end{figure*}
\begin{table}[!t]
    \centering
    \caption{Ablation study of each component in PAMT using DTFD with CNN and ViT backbones on the Camelyon16 dataset. `+' denotes additional trained parameters beyond the common MIL classifier.}
    \renewcommand\arraystretch{1.2}
    \setlength{\tabcolsep}{7pt}{
    \begin{tabular}{cccccc}
        \toprule
        Base & RPS & PVP & AMT & AUC & Parameters (M)\\
        \midrule
        \rowcolor{whitesmoke} 
        \multicolumn{6}{c}{ResNet-50} \\
        \midrule
        \checkmark &&&& 0.910 & -- \\
        \checkmark &\checkmark&&& 0.916 & +0 \\
        \checkmark &\checkmark&\checkmark&& 0.922 & +0.118 \\
        \checkmark &\checkmark&&\checkmark& 0.924 & +0.393 \\
        \checkmark &\checkmark&\checkmark&\checkmark& \textbf{0.929} & +0.511 \\
        \midrule
        \rowcolor{whitesmoke} 
        \multicolumn{6}{c}{ViT-b-32} \\
        \midrule
        \checkmark &  &  &  & 0.887 & -- \\
        \checkmark & \checkmark &  &  & 0.890 & +0\\
        \checkmark & \checkmark & \checkmark &  & 0.894 & +0.011 \\
        \checkmark & \checkmark &  & \checkmark & 0.917 & +0.031 \\
        \checkmark & \checkmark & \checkmark & \checkmark & \textbf{0.936} & +0.042 \\
        \bottomrule
    \end{tabular}}
    \label{ablation1_component}%
\end{table}

\begin{table}[h]
\centering
\small
\caption{Comparison with alternative parameter-efficient fine-tuning strategies on the Camelyon16 dataset.}
\renewcommand\arraystretch{1.2}
\setlength{\tabcolsep}{7pt}
\begin{tabular}{lcccc}
\toprule
Method & AUC & F1 & Acc & Param.(M) \\
\midrule
VPT-s & 0.926$_{0.015}$ & 0.807$_{0.056}$ & 0.826$_{0.044}$ & 0.59 \\
VPT-d & 0.927$_{0.022}$ & 0.868$_{0.036}$ & 0.872$_{0.033}$ & 0.60 \\
LoRA & 0.944$_{0.003}$ & 0.877$_{0.019}$ & 0.880$_{0.020}$ & 0.60 \\
AdaLoRA & 0.944$_{0.003}$ & 0.877$_{0.019}$ & 0.880$_{0.020}$ & 0.60 \\
IA\textsuperscript{3} & 0.950$_{0.005}$ & 0.892$_{0.010}$ & 0.896$_{0.008}$ & 0.53 \\
MeLo & 0.941$_{0.005}$ & 0.874$_{0.020}$ & 0.880$_{0.020}$ & 0.60 \\
AdaptF & 0.949$_{0.004}$ & 0.895$_{0.015}$ & 0.899$_{0.015}$ & 0.66 \\
SSF & 0.952$_{0.005}$ & \textbf{0.901}$_{0.010}$ & 0.905$_{0.010}$ & 0.78 \\
PAMT & \textbf{0.959}$_{0.006}$ & 0.887$_{0.009}$ & \textbf{0.915}$_{0.008}$ & 0.61 \\
\bottomrule
\end{tabular}
\label{tab:ablation1_component}
\end{table}
\subsubsection{Effectiveness of Each Component.}
We conduct a component ablation study to evaluate the effectiveness of representative patch sampling (RPS), prototypical visual prompt (PVP), and adaptive model transformation (AMT) under both CNN and Transformer backbones. As shown in Table~\ref{ablation1_component}, with the ResNet-50 backbone, the baseline DTFD model achieves an AUC of 0.910 on Camelyon16. After applying RPS, the AUC increases to 0.916, while PVP and AMT further improve the AUC to 0.922 and 0.924, respectively. Integrating all three techniques yields an AUC of 0.929. We further observe the same favorable trend on the ViT-b-32 backbone, where the complete PAMT configuration reaches an AUC of 0.936. These results demonstrate that the three components contribute complementary gains across both CNN and Transformer architectures.

\subsubsection{Comparison with Other Tuning Strategies.}
In Table 6, we compare our method with several representative tuning strategies on the Camelyon16 dataset, including Fully-tuning, Linear probing, Visual Prompt Tuning (VPT)~\citep{jia2022visual}, LoRA~\citep{hu2021lora}, AdaLoRA~\citep{zhang2023adaptive}, IA$^3$~\citep{liu2022fewshot}, MeLo~\citep{yu2024melo}, AdaptFormer~\citep{chen2022adaptformer}, and SSF~\citep{lian2022scaling}. We report AUC, Accuracy, and F1 score together to provide a more comprehensive evaluation of classification quality. While generic PEFT methods such as VPT, LoRA, AdaLoRA, IA$^3$, MeLo, AdaptFormer, and SSF reduce trainable parameters and partially alleviate overfitting, they still apply a largely uniform adaptation mechanism across highly heterogeneous histopathology patches. In contrast, PAMT combines representative patch sampling, cluster-aware visual prompting, and lightweight model adaptation, which leads to better overall performance (e.g., highest AUC and Accuracy) without incurring substantial parameter overhead. These results validate the necessity of explicitly modeling histopathological heterogeneity rather than relying only on generic PEFT modules.
\begin{table}[h]
\centering
\caption{Sensitivity analysis of the number of selected patches $K$ in RPS within the PAMT framework on the Camelyon16 dataset.} 
\renewcommand\arraystretch{1.2}
\setlength{\tabcolsep}{6pt} 
\resizebox{\linewidth}{!}{
\begin{tabular}{cccccc}
\toprule
\shortstack{$K$ (Patches\\/WSI)} & AUC & Acc & F1 & \shortstack{Train Peak\\ Mem. (GB)}  & \shortstack{Train Time\\/Epoch (min)} \\
\midrule
32 & 0.918 & 0.853 & 0.842 & 3.07 & 0.28 \\
64 & 0.925 & 0.871 & 0.860 & 5.99 & 0.40 \\
128 & \textbf{0.932} & \textbf{0.881} & \textbf{0.875} & 11.83 & 0.72 \\
256 & 0.930 & 0.876 & 0.870 & 23.15 & 1.57 \\
512 & 0.924 & 0.864 & 0.855 & 45.78 & 2.86 \\
\bottomrule
\end{tabular}
}
\label{tab:rps_topk}
\end{table}

\begin{table}[h]
\centering
\caption{Comparison of different patch selection strategies within the PAMT framework on the Camelyon16 dataset. All methods select $K=128$ patches per WSI and are averaged over three random seeds.} 
\renewcommand\arraystretch{1.2}
\setlength{\tabcolsep}{6pt} 
\resizebox{\linewidth}{!}{
\begin{tabular}{lccc}
\toprule
Sampling Strategy & AUC & Acc & F1 \\
\midrule
Mean-pooling & 0.610 $\pm$ 0.053 & 0.633 $\pm$ 0.013 & 0.470 $\pm$ 0.097 \\
Uncertainty-aware sampling~\citeyear{aljuhani2022uncertainty} & 0.569 $\pm$ 0.002 & 0.612 $\pm$ 0.019 & 0.556 $\pm$ 0.011 \\
Clustering-based sampling~\citeyear{yang2022remix} & 0.729 $\pm$ 0.002 & 0.703 $\pm$ 0.124 & 0.669 $\pm$ 0.103 \\
Ours (RPS) & \textbf{0.920 $\pm$ 0.004} & \textbf{0.837 $\pm$ 0.016} & \textbf{0.816 $\pm$ 0.020} \\
\bottomrule
\end{tabular}
}
\label{tab:rps_comparison}
\end{table}
\subsubsection{Analysis of Representative Patch Sampling Strategies.}
To rigorously validate the robustness and necessity of the Representative Patch Sampling (RPS) module, we conduct two additional analyses on Camelyon16. First, we evaluate different values of $K$, the number of selected patches per WSI. As shown in Table~\ref{tab:rps_topk}, performance improves as $K$ increases from 32 to 128 and peaks at this value, while larger $K$ values introduce additional memory and training-time costs without yielding further gains. This validates our choice of $K=128$ as a practical balance between diagnostic coverage and computational feasibility. Second, under the same budget of $K=128$, we compare attention-based RPS with alternative patch selection strategies, including diversity-oriented clustering-based sampling~\cite{yang2022remix}, uncertainty-aware sampling~\cite{aljuhani2022uncertainty}, and mean-pooling. As shown in Table~\ref{tab:rps_comparison}, attention-guided sampling dominates all alternative selection strategies, yielding a substantial 26\% to 62\% improvement in AUC (e.g., 0.920 vs. 0.729 for the second-best clustering-based sampling). This comprehensively demonstrates that RPS is a highly stable and effective mechanism for retaining crucial diagnostic regions when drastically reducing WSI bag size before end-to-end fine-tuning with PVP and AMT.

\subsubsection{Analysis under Limited Data.}
To further evaluate the robustness of our method, we explore more challenging scenarios with varying proportions of the training set within the TCGA-NSCLC dataset.
As depicted in Figure~\ref{fig:wsi_num}, our method yields consistent superiority over the LP and full-tuning methods.
Notably, employing merely 75\% of the training data, our approach surpasses the LP performance with the whole dataset, underscoring the efficacy of our method in extracting and leveraging meaningful patterns from limited data and the potential for application in scenarios where data acquisition is challenging.
In contrast, the full-tuning approach consistently underperforms across all data proportions, which may initially indicate a propensity for overfitting. 
However, its performance gap narrows with the enlargement of the dataset. 
This trend implies that while larger datasets can mitigate overfitting to an extent, the critical factor lies in the strategic utilization of the available data, a domain where our proposed method excels, ensuring robust performance even with smaller datasets. 
This proficiency in data utilization highlights the transferability of our method, advocating its suitability for effective deployment in resource-constrained environments.
\begin{table}[htbp]
\caption{Analysis of adapter position across layers.}
\centering
\label{tab:adapter_postion}%
\renewcommand\arraystretch{1.2}
\setlength{\tabcolsep}{8pt}{
\begin{tabular}{cccccc}
\toprule
\multirow{3}{*}{Location} & Layer1 & \checkmark     & \checkmark     &   &       \\
                          & Layer2 & \checkmark     &       & \checkmark &       \\
                          & Layer3 & \checkmark     &       &   & \checkmark     \\ 
\midrule
AUC                       &        & 0.918 & 0.916 &0.919   & \textbf{0.921} \\ 
F1            &     & 0.820 & 0.815 & 0.823 & \textbf{0.828} \\
Accuracy      &     & 0.860 & 0.858 & 0.861 & \textbf{0.866} \\
\bottomrule
\end{tabular}}
\end{table}
\subsubsection{Analysis of Hyper-parameter Sensitivity.}
\label{sec:hyperparameter}
We conduct an ablation study to evaluate the hyper-parameter sensitivity of our method, using AB-MIL on the Camelyon16 dataset.
Each experiment is averaged across three runs.
\begin{figure}[thbp]
    \centering
    \includegraphics[width=\linewidth]{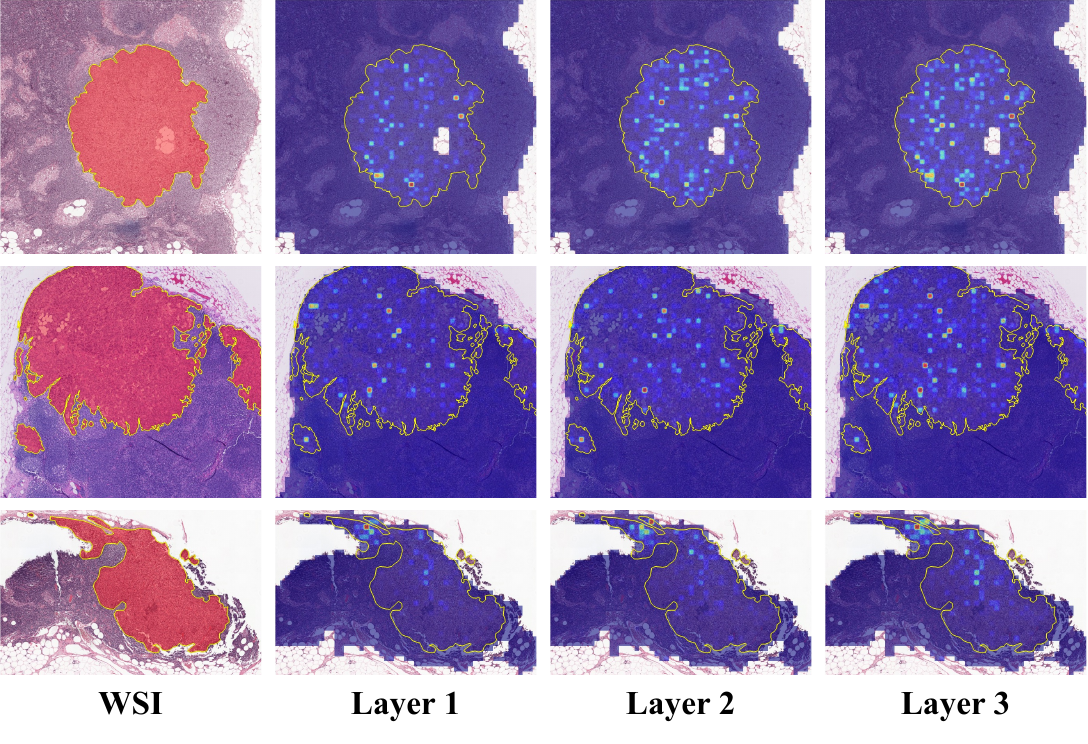}
    \caption{Attention map comparison across different layers of the model. (left) Layer 1; (middle) Layer 2; (right) Layer 3.}
	\label{fig:adapter}
\end{figure}
\myparagraph{Adapter position.}
In Table~\ref{tab:adapter_postion}, we conduct an ablation study to evaluate the effectiveness of adapter across different layers using AB-MIL on Camelyon16. 
The results highlight that placement in the third layer, closer to the network's output, significantly enhances model accuracy, as indicated by the highest AUC of 0.921, and underscores the strategic impact of adapter block depth on capturing complex histopathological features.
We also visualized the attention maps across different layers of the model to understand how the placement of the AMT modules affects feature learning. As shown in Figure~\ref{fig:adapter}, we observe that placing AMT modules at deeper layers (closer to the output) results in more focused attention on diagnostically relevant regions, while earlier placements lead to more diffuse attention patterns. This suggests that deeper AMT modules are more effective at correcting domain-specific feature representations, further supporting our choice of adapter placement.

\myparagraph{Number of adapter blocks.}
In Figure~\ref{fig:adapter_num}, we conduct experiments with different numbers of adapter blocks in the last layer of ResNet-50.
Results demonstrate that the AUC increases with the number of adapter blocks, reaching a peak at 3 blocks, and then slightly decreases with more blocks.
This indicates that 3 adapter blocks are sufficient to capture the complex histopathological features, while more blocks may lead to redundancy.

\myparagraph{Number of clusters in prototypical visual prompt.}
We apply different numbers of clusters in the prototypical visual prompt (PVP) in Figure~\ref{fig:cluster_num}.
Results demonstrate that the AUC increases with the number of clusters, reaching a peak at 4 clusters, and then slightly decreases with more clusters.
This number of clusters allows the model to effectively represent the variety of histopathological features, enhancing diagnostic relevance without succumbing to over-complexity.

\myparagraph{Padding size in prototypical visual prompt.}
We evaluate the impact of different padding sizes in PVP in Figure~\ref{fig:padding_size}.
Results show that a padding size of 10 achieves the highest AUC, signifying an ideal spatial dimension for capturing salient features while limiting extraneous details.
Increases beyond this size do not yield performance gains, suggesting that larger padding may introduce noise or irrelevant information, thereby diluting the model's focus on critical histopathological patterns.

\subsubsection{Visualization of Cluster Centers for PVP.}
In Figure~\ref{fig_ablation_visual}, we visualize the cluster centers obtained from the representative patch sampling strategy, which discerned four salient histopathological patterns matched with distinct padding prompts.
Results demonstrate a spectrum of tissue characteristics, from sparse inflammatory infiltrates to areas of dense nuclear aggregation, which are likely to carry diagnostic relevance.
This visualization demonstrates the clustering process's capacity to isolate and highlight significant histopathological features within the dataset.
\begin{table}[thbp]
\centering
\caption{Cross-center independent validation results in AUC. Models trained on Camelyon16 are evaluated directly on the Camelyon17 dataset.}
\setlength{\tabcolsep}{4pt}
\begin{tabular}{lcccc}
\toprule
Method & Backbone & \shortstack{In-domain\\ Camelyon16} & \shortstack{Cross-center\\ Camelyon17} & Drop \\
\midrule
DTFD-MIL  & ResNet-50 & 0.910$_{0.011}$ & 0.840$_{0.001}$ & -7.0 \\
PAMT & ResNet-50 & 0.929$_{0.010}$ & 0.871$_{0.005}$ & -5.8 \\
DTFD-MIL  & ViT-b-32 & 0.946$_{0.009}$ & 0.841$_{0.005}$ & -10.5 \\
PAMT  & ViT-b-32 & 0.959$_{0.006}$ & 0.881$_{0.006}$ & -7.8 \\
\bottomrule
\end{tabular}
\label{tab:cross_center_validation}
\end{table}
\subsubsection{Cross-Center Independent Validation.}
To evaluate the generalization capability and robustness of the complete PAMT framework under cross-center and cross-scanner domain shifts, we conducted an external independent validation. Models trained exclusively on the Camelyon16 dataset were evaluated directly on the independent, multi-center Camelyon17 cohort~\citep{bandi2018detection} without any further fine-tuning. The Camelyon17 dataset introduces significant variability in tissue preparation, staining protocols, and slide scanners across five different medical centers. As detailed in Table~\ref{tab:cross_center_validation}, for both backbones, PAMT achieves higher in-domain performance than the baseline DTFD-MIL model (0.929 vs. 0.910 for ResNet-50; 0.959 vs. 0.946 for PLIP) and exhibits more robust cross-center generalization. Specifically, when transferred to the external cohort, the performance drop is reduced from 7.0\% to 5.8\% for ResNet-50, and from 10.5\% to 7.8\% for PLIP. This demonstrates that the synergistic integration of Representative Patch Sampling (RPS), Prototypical Visual Prompts (PVP), and Adaptive Model Transformation (AMT) concentrates on the most diagnostically discriminative tissue regions and adapts the backbone without overfitting to the source-center stain and scanner distribution, enabling better transferability to a new center.
\begin{table}[thbp]
\centering
\small
\caption{System-level efficiency comparison on the Camelyon16 dataset (tested on a single NVIDIA H100 NVL 94 GB GPU).}
\renewcommand\arraystretch{1.2}
\setlength{\tabcolsep}{4pt} % Adjust column spacing
\begin{tabular}{lcccc}
\toprule
Method & \shortstack{Train\\ Patches} & \shortstack{Train\\ Param (M)} & \shortstack{Train Peak\\ Mem. (GB)} & \shortstack{Train\\ Latency (s)} \\
\midrule
Frozen-encoder & 1075 & 1.31 & 0.80 & 0.004 \\
Full Fine-Tuning & 1075 & 24.82 & 86.51 & 0.616 \\
PAMT (Ours) & 128  & 2.65 & 11.87 & 0.058 \\
\bottomrule
\end{tabular}
\label{tab:system_efficiency}
\end{table}
\subsubsection{Analysis of System-Level Computational Efficiency.}
While parameter counts provide a proxy for model size, we further evaluate the system-level computational efficiency of the PAMT framework by analyzing GPU memory consumption, parameter scalability, and training latency. As detailed in Table~\ref{tab:system_efficiency}, full-network fine-tuning is highly memory-intensive and computationally prohibitive due to the necessity of backpropagating gradients through the entire backbone for large WSI bags (e.g., 1075 patches). Specifically, full fine-tuning demands up to 86.51 GB of peak memory and 24.82 M trainable parameters, requiring high-end data center GPUs (e.g., NVIDIA H100 94 GB). In contrast, PAMT leverages Representative Patch Sampling (RPS) to reduce the bag size to $K=128$ patches per slide. Although the PAMT pipeline introduces a minor preprocessing overhead (training the initial MIL classifier and performing k-means clustering on the sampled top-K features), this cost is heavily offset during the tuning phase. By freezing the pre-trained feature extractor and updating only the visual prompts, adapter blocks, and the lightweight MIL classifier (requiring only 2.65 M trainable parameters), PAMT drastically reduces GPU memory usage by 86.3\% (from 86.51 GB to 11.87 GB) compared to full fine-tuning. This exceptional memory efficiency allows PAMT to be seamlessly trained on a standard research-grade GPU such as the NVIDIA GeForce RTX 4090 (24 GB). Furthermore, PAMT reduces the training latency to 0.058 seconds per WSI, which is an order of magnitude faster than full fine-tuning (0.616 seconds). Although slightly higher than the frozen-feature DTFD-MIL baseline (0.004 seconds) due to the addition of adapter modules, the training latency remains highly comparable and well within practical limits for clinical workflows, confirming that PAMT achieves state-of-the-art accuracy without compromising system-level efficiency.
\begin{table}[htbp]
\centering
\small
\caption{Quantitative evaluation of PVP cluster quality on the Camelyon16 dataset ($C=4$). ``Full'' refers to clustering all available patches, while ``RPS'' restricts clustering to the RPS-selected subset ($K=128$).}
\renewcommand\arraystretch{1.2}
\setlength{\tabcolsep}{2pt} 
\begin{tabular}{lccccc}
\toprule
Backbone & \shortstack{Input\\ Source} & \shortstack{Silhouette\\ Score $\uparrow$} & \shortstack{Intra-cluster\\Mean Dist $\downarrow$} & \shortstack{Inter-cluster\\Mean Dist $\uparrow$} & \shortstack{Inter/Intra\\Ratio $\uparrow$} \\
\midrule
\multirow{2}{*}{ResNet-50} & Full & 0.165 & 0.664 & 0.936 & 1.41$\times$ \\
 & RPS & \textbf{0.283} & \textbf{0.62} & \textbf{1.480} & \textbf{2.40$\times$} \\
\midrule
\multirow{2}{*}{ViT-b-32} & Full & 0.162 & 3.621 & 4.331 & 1.20$\times$ \\
 & RPS & \textbf{0.203} & \textbf{3.40} & \textbf{6.460} & \textbf{1.90$\times$} \\
\bottomrule
\end{tabular}
\label{tab:pvp_cluster_quality}
\end{table}
\begin{figure}[thbp]
    \centering
    \includegraphics[width=\linewidth]{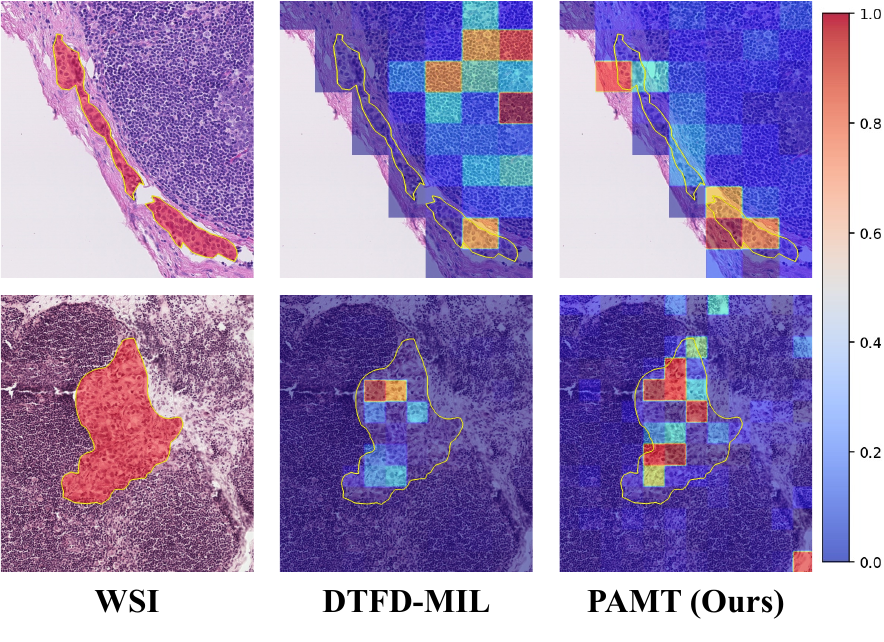}
    \caption{Comparison of attention heatmaps on the Camelyon16 dataset. (left) Original H\&E whole slide image with tumor regions annotated. (middle) Attention heatmap generated by the baseline DTFD-MIL model. (right) Attention heatmap generated by the PAMT framework.}
	\label{fig:attention_heatmaps}
\end{figure}
\begin{figure}[thbp]
    \centering
    \includegraphics[width=\linewidth]{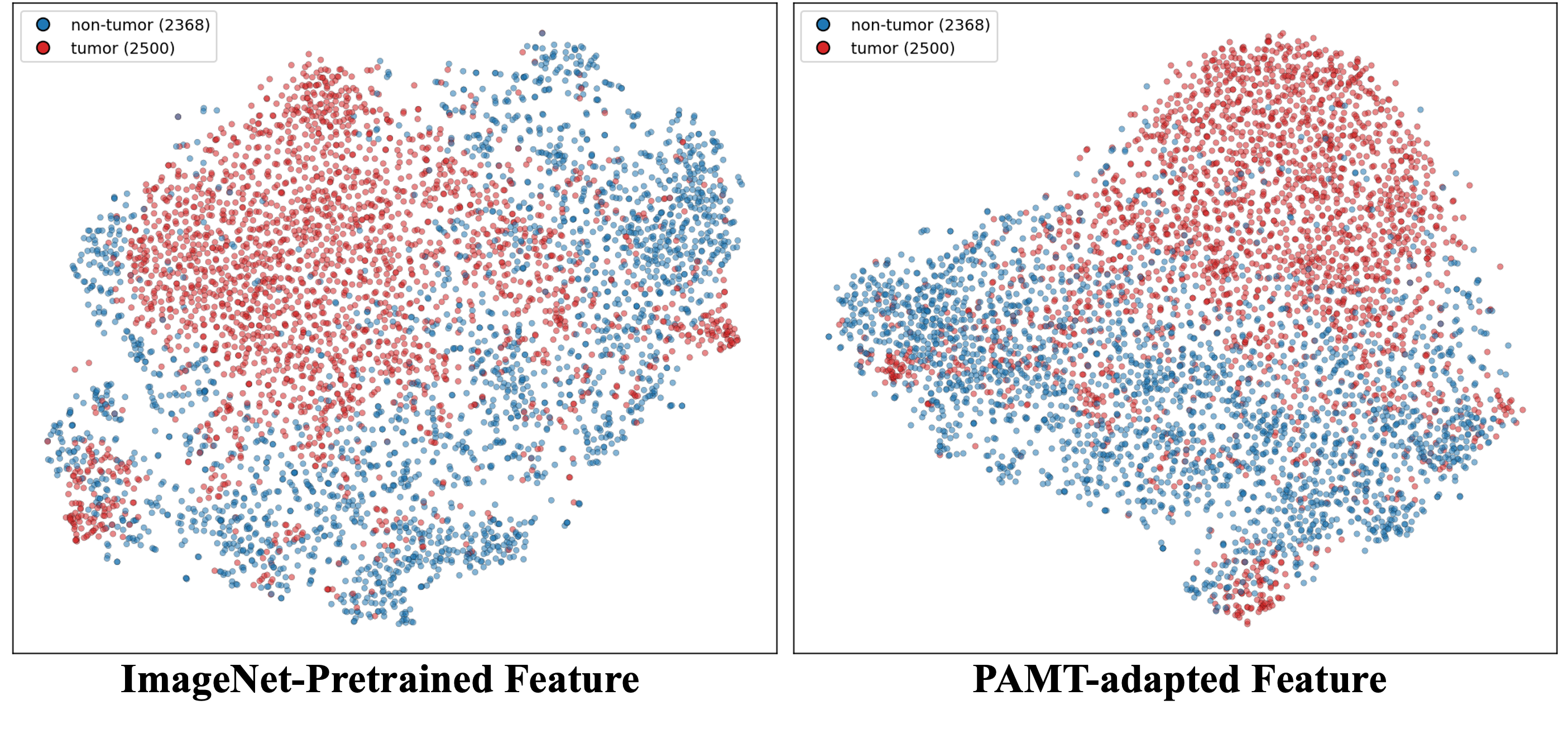}
    \caption{t-SNE visualization of WSI-level features before and after AMT. (a) Frozen ImageNet features. (b) PAMT-adapted features.}
	\label{fig:tsne_adapter}
\end{figure}
\subsubsection{Interpretability and Feature Alignment Analysis.}
To provide quantitative and deeper interpretability of the PAMT framework, we analyzed the feature alignment, attention distributions, and clustering quality. First, we quantitatively assessed the PVP clustering process using the Silhouette score and the ratio of inter-to-intra cluster distances. As shown in Table~\ref{tab:pvp_cluster_quality}, on the Camelyon16 dataset with the ResNet-50 backbone, clustering on the $K=128$ representative patches yields a Silhouette score of 0.283 and an inter/intra distance ratio of 2.4$\times$. For the PLIP backbone, the Silhouette score is 0.203 with a ratio of 1.9$\times$. In contrast, when clustering is applied to the full feature bag (unfiltered WSI patches), the Silhouette score drops significantly to 0.165 (ResNet-50) and 0.162 (PLIP), and the inter/intra ratio decreases as well. This robustly demonstrates that the Representative Patch Sampling (RPS) module is more than just a computational heuristic; it actively filters out noisy background patches to select a semantically cleaner and more separable subset of histopathological phenotypes, thereby ensuring high intra-cluster consistency for prompt initialization. Furthermore, as illustrated in Figure~\ref{fig:attention_heatmaps}, we compare the attention heatmaps generated by the baseline DTFD-MIL model against those from the PAMT framework. The visualization confirms that PAMT adaptively reshapes model attention, exhibiting a significantly higher concentration of high-attention scores on confirmed tumor regions while suppressing noise from non-diagnostic stroma and artifacts. Finally, t-SNE visualization of the WSI-level feature representations in Figure~\ref{fig:tsne_adapter} demonstrates that the Adaptive Model Transformation (AMT) modules successfully correct the initial domain shift. The ImageNet-pretrained ResNet features exhibit poor class separability, whereas the PAMT-adapted features form distinct, well-separated clusters corresponding to the clinical diagnostic categories, confirming that the framework learns meaningful, domain-specific corrections at the feature level.
\begin{figure}[thbp]
    \centering
    \includegraphics[width=\linewidth]{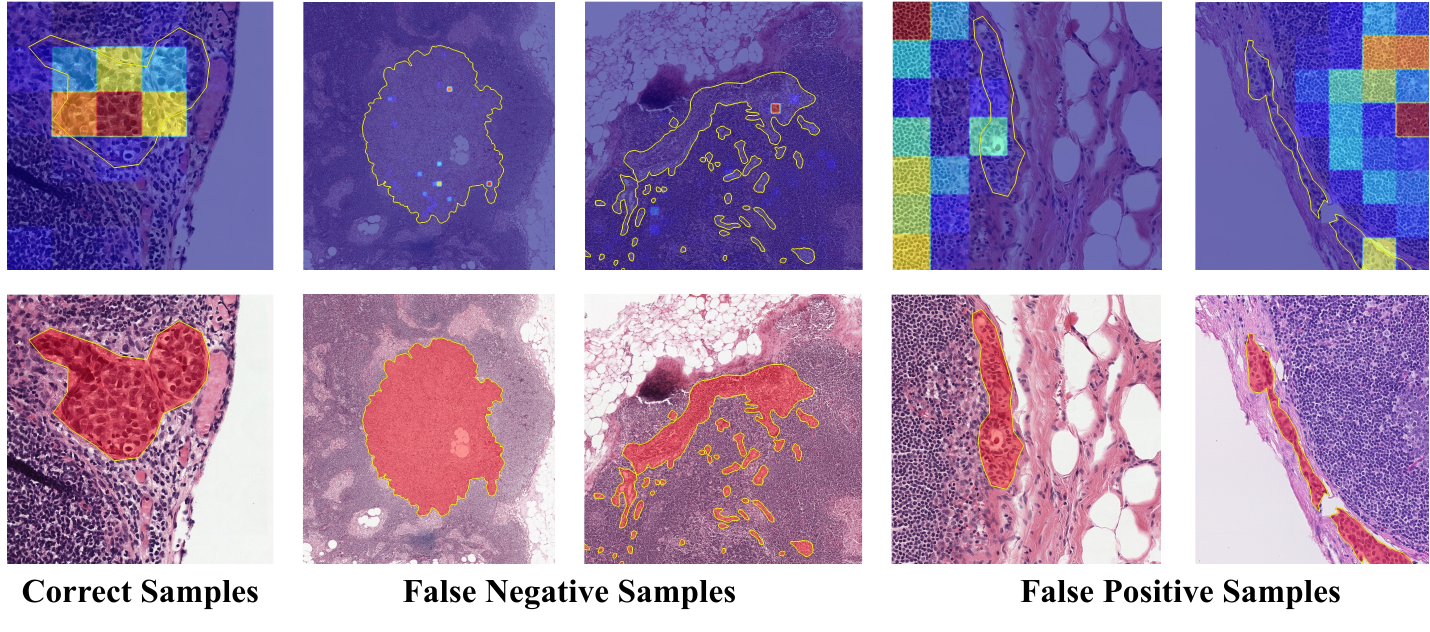}
    \caption{Representative failure cases of the PAMT framework. (column 1) A True Positive (TP) case correctly identified by the model. (columns 2--3) False negatives that were assigned a low initial attention score and subsequently excluded by the RPS module. (columns 4--5) False positives that disrupted the PVP feature clustering and misled the model's adapted attention.}
    \label{fig:error}
\end{figure}
\subsubsection{Error Analysis and Failure Cases.}
While PAMT significantly improves overall classification performance, analyzing its failure cases provides important insights into remaining challenges. Figure~\ref{fig:error} illustrates two primary scenarios where the framework misclassifies whole slide images. The first scenario (in columns 2--3) involves extremely sparse micro-metastases. Because the Representative Patch Sampling (RPS) module aggressively reduces the WSI to the top-$K$ patches based on baseline attention scores, highly localized anomalies in otherwise benign tissue can occasionally be discarded before reaching the fine-tuning stage, leading to false negatives. The second scenario (in columns 4--5) involves severe out-of-distribution staining artifacts (e.g., massive dye pooling or tissue folding). In these cases, the offline k-means clustering may group these artifacts into a dominant cluster, causing the Prototypical Visual Prompts (PVP) to over-attend to technical noise rather than underlying morphology. Future iterations of PAMT will explore dynamic, learnable sampling thresholds and stain-invariant prompt initialization to mitigate these specific failure modes.
% -----------------------------------------------------
\section{Limitations}
\label{sec:limitations}
The current implementation of the \texttt{PAMT} framework has several limitations that should be considered for future research:
\begin{itemize}
    \item \textbf{Patch Selection:} The framework relies on a patch selection strategy that may not always yield the most informative patches for classification. The choice of patches can significantly impact the performance of the model, and the current implementation does not adaptively select patches based on their relevance to the classification task.
    \item \textbf{Model Complexity:} The framework uses a complex model architecture that may require significant computational resources and time for training. This can be a barrier for users with limited computational capabilities or those who need to quickly iterate on model designs.
    \item \textbf{Generalization:} While the framework has shown promising results on 14 datasets in total, including 13 primary datasets and the Camelyon17 external validation cohort, its generalization capabilities to other diverse types of WSIs (e.g., rare morphological subtypes) or entirely different clinical endpoints are not fully evaluated. Future work will investigate the scalability of PAMT across broader multi-modal prognostic tasks.
\end{itemize}
% -----------------------------------------------------
\section{Conclusion}
In this study, we present \texttt{PAMT}, a Prompt-guided Adaptive Model Transformation framework designed to advance pathology image classification.
\texttt{PAMT} directly confronts the domain shift and task heterogeneity between pre-training datasets and histopathological image analysis, which often results in suboptimal feature representations.
To mitigate these challenges, \texttt{PAMT} adopts a dual-pronged strategy encompassing both input and model reprogramming to systematically refine the data representation and the foundational model architecture.
During the input reprogramming phase, we introduce Representative Patch Sampling (RPS) and Prototypical Visual Prompt (PVP), enabling the encapsulation of intricate pathological feature distributions across diverse histopathological slides.
In the model reprogramming stage, we develop Adaptive Model Transformation (AMT) by incorporating scalable adapter modules into pre-trained feature extractors, thereby facilitating an efficient transfer of knowledge from general-purpose models to domain-specific downstream tasks.
Extensive validation on 14 publicly available datasets, employing six state-of-the-art pathology image classifiers, substantiates that \texttt{PAMT} substantially enhances pathology image classification at both the whole-slide and patch levels, thus demonstrating its efficacy and computational efficiency.
Future work will focus on designing more sophisticated visual prompts and adapter modules, with particular emphasis on scalability, flexibility, and robustness.
% -----------------------------------------------------
\section*{Declaration of generative AI and AI-assisted technologies in the writing process}
During preparing this work, the authors used ChatGPT 4.0 in order to improve language and readability. After using this tool, the authors reviewed and edited the content as needed and take full responsibility for the content of the publication.

\section*{Declaration of Competing Interest}
None declared.

\section*{Acknowledgement}
This work was supported by the National Key R\&D Program of China (Project No. 2023YFE0204000), National Natural Science Foundation of China (No. 62202403), Innovation and Technology Commission (Project No. GHP/006/22GD and ITCPD/17-9), Research Grants Council of the Hong Kong Special Administrative Region, China (Project No: T45-401/22-N) and Shenzhen Science and Technology Innovation Committee Fund (Project No. KCXFZ20230731094059008).  
% -----------------------------------------------------
\bibliographystyle{model2-names}
\biboptions{authoryear}
\bibliography{refs}
% -----------------------------------------------------
\end{document}